# A LARGE LANGUAGE MODEL FOR DISASTER STRUCTURAL RECONNAISSANCE SUMMARIZATION

Y. Gao[1], G. Zhou[2] & K. M. Mosalam[3]

[1] University of California Berkeley, Berkeley, United States, gaoyuqing@berkeley.edu Current address: Tongji University, Shanghai, China

[2] University of California Berkeley, Berkeley, United States, guanren_zhou@berkeley.edu

[3] University of California Berkeley, Berkeley, United States, mosalam@berkeley.edu

**Abstract**: *With recent advances in Computer Vision (CV) and Deep Learning (DL), Artificial Intelligence (AI)-aided vision-based Structural Health Monitoring (SHM) has emerged as a promising technique for monitoring and rapidly assessing the health of structures. By using cameras and CV algorithms to analyze images and videos of a structure, vision-based SHM can detect subtle changes in appearance that may indicate damage or deterioration. Applying DL to vision-based SHM has shown great potential in improving the accuracy and efficiency of the process, as DL models can learn to classify and localize visual patterns associated with different types of structural damage and efficiently making use of big data in the form of large number of images and videos. However, the DL models in previous works only generate discrete outputs, e.g., damage class labels and damage region's coordinates, and the engineers still need to put extra efforts to re-organize and analyze these results for further evaluation and decision-making. In late 2022, Large Language Models (LLMs), such as "Chat Generative Pre-trained Transformer" (ChatGPT), have become popular in various fields, which brings new insights into the AI-aided vision-based SHM. In this study, a novel LLM-based Disaster Reconnaissance Summarization (LLM-DRS) framework is proposed. It first introduces a standard reconnaissance plan, where the collection of both vision data (e.g., images) and corresponding metadata of structures should follow a well-designed on-site investigation plan. Subsequently, text-based metadata and image-based vision data are further processed and matched to the designated format into one file, where the well-trained Deep Convolution Neural Networks from the PEER Hub ImageNet ($\phi$-Net) extract the key text-based attributes, such as damage state, material type, and damage levels, from these images. Finally, feeding all data into a GPT model with a set of carefully crafted prompts, the LLM-DRS generates the summary report for: (1) an individual structure, or (2) the affected regions based on the attributes and metadata from all investigated structures within the region. It is shown that the integration of LLMs in vision-based SHM, especially for rapid post-disaster reconnaissance, achieves promising results indicating the great potential to achieve resilient built environment through effective reconnaissance.*

## 1 Motivations

With the rapid development of Artificial Intelligence (AI) technologies, many areas have benefited from AI's automation, effectiveness, efficiency, and economics. In recent years, disasters, caused by natural hazards, e.g., earthquakes, tsunamis, and hurricanes, occurred frequently worldwide. After the occurrence of a disaster, engineers and investigation teams often go to the affected site as soon as possible to conduct reconnaissance and collect perishable data to provide rapid assessment of the built environment, in order to assist in post-disaster rescue and reduce consequent losses. In the past few years, many efforts (Gao & Mosalam, 2018; 2020; 2023) have been performed in rapid assessment and structural health monitoring (SHM) using Computer Vision (CV) and Deep Learning (DL) algorithms to analyze images and videos of a structure to detect subtle changes in appearance that may indicate damage or deterioration. These AI technologies have shown great potential in improving the accuracy and efficiency of the process, as DL models can learn to classify and localize visual patterns associated with different types of structural damage.

Previous works are more task-specific and focus on generating discrete outputs, e.g., damage class labels, damage region's coordinates, and damage areas for classification, localization, and segmentation tasks. The engineers still need to put extra effort to re-organize and analyze these results for further evaluation and decision-making. Furthermore, a large amount of metadata is collected during reconnaissance, such as geolocation (latitude and longitude) of the investigated damaged building, earthquake magnitude, building types, and overall ratings. These metadata can directly contribute to the reconnaissance report, but they are not fully exploited in the previous AI-aided SHM studies, where AI Generative Content (AIGC) techniques hold promise for addressing this gap.

In this paper, the major objective is to integrate AI technologies to rapidly generate structural health assessment reports for a single structure or a region containing multiple structures after a disaster. This is conducted by providing a rapid summarization using vision data and metadata collected during the reconnaissance effort. To tackle this problem, a multi-modality framework integrating technologies of Convolution Neural Network (CNN), Natural Language Processing (NLP), and Large Language Model (LLM) is proposed, named LLM-based Disaster Reconnaissance Summarization (LLM-DRS) framework. The major computational novelties and contributions of this LLM-DRS framework are:

1) It first utilizes the well-trained CNN models to extract key attributes from collected images, such as structural or component type, damage state, and damage level.

2) It adopts "prompt engineering", a process of carefully constructing prompts or inputs for AI models to enhance their performance on specific generation tasks, to design a set of rules and text templates for AI to generate assessment reports.

3) It implements LLMs, such as GPT-4, to link the structural attributes extracted from step (1) and extra collected metadata through the prompt design in step (2), and to finally generate professional reconnaissance reports for a single structure or a region.

## 2   Related Works

The research on NLP has a long history. It is generally believed that it has been inspired by the famous Turing Test proposed by Alan Turing in 1950 (Turing, 2008) at the time when researchers have been developing methods that machines can understand and interact with humans with natural languages. The development of NLP can be considered to have gone through the following four stages: rule-based methods, statistics-based methods, machine learning-based methods, and deep learning-based methods. Since 2017, the introduction of the Transformer (Vaswani et al., 2017), distinguished by its attention mechanism, parallel processing capabilities, and the capacity to capture long-range dependencies within natural language sentences, has brought about a noteworthy improvement in the field of NLP. The architecture of the Transformer, characterized by its flexibility, scalability, and a vast number of learnable parameters, paved the way for the more effective use of the pre-training and fine-tuning paradigm in NLP tasks. Based on the success of the Transformer, several variants were proposed and have been proven to be powerful, where the Bidirectional Encoder Representations from Transformers (BERT) (Devlin et al., 2018) and Generative Pre-trained Transformer (GPT) (Radford et al., 2018; Radford et al., 2019; Brown et al., 2020) are two of the most notable ones. Both BERT and GPT are pre-trained on vast corpus utilizing unsupervised learning methods. BERT, derived from the encoder of the Transformer, leverages bidirectional context and has shown exemplary performance on several downstream tasks requiring contextual understanding, such as text classification. On the other hand, GPT, inspired by the Transformer's decoder, excels in generative tasks as a Language Model (LM), with ChatGPT being one of its well-known fine-tuned incarnations.

In recent years, research on AI-aided SHM and post-disaster reconnaissance has mainly focused on adopting CV methods, especially CNN, to automate some tasks that require manual labor and key information about the structure's health condition from photographs obtained during the field investigation. Gao and Mosalam (2018) proposed the concept of Structural ImageNet and built one of the largest open-source image datasets in the SHM area, namely, PEER Hub ImageNet ($\phi$-Net), and used it as a benchmark for vision-based SHM task studies. Recently, Gao and Mosalam (2020; 2023) further explored the power of the Transformer and developed a holistic framework, namely, Multi-Attribute Multi-Task Transformer (MAMT2) framework to simultaneously perform structural image classification, localization, and segmentation tasks.

Compared with the booming development in vision-based SHM, the NLP-based approaches received little attention in SHM. Some relevant works focus on using NLP methods to extract relevant information about disasters from social media posts (e.g., X, previous Twitter). Tsai et al. (2020) developed an approach to estimate the recovery time of an earthquake event by analyzing the occurrence of social media posts on specific topics over time, which is based on an assumption that posts on some event-related topics appear more frequently at the beginning of the event, and reduce back to normal level when the community has been recovered. Fersini et al. (2017) designed a support system that uses a combination of several models to analyze the tweets from 4 dimensions (What, Who, Where, and When) to address early warning signals and support the final decision-making, which showed the great potential of NLP methods in the management of earthquake events.

Beyond the conventional NLP technologies, LLM has become the most state-of-the-art realization of NLP and has achieved great success in the past few years. However, to the best of the authors' knowledge, it has very limited exploration in the domain of SHM. Only very limited early-stage studies in other Civil Engineering-related fields, such as Engineering Management, are reported. For instance, Prieto et al. (2023) tested the applicability of ChatGPT in construction project scheduling by conducting an experiment and concluded that the overall performance of ChatGPT is reasonable and most of the engineers who joined the experiment gave positive feedback to the interaction experience. Moreover, the utility of LLMs extends beyond the conventional "pre-train" and "fine-tune" paradigm, which typically relies on additional supervised data for specific tasks. LLMs can be adeptly leveraged for downstream tasks by using carefully crafted prompts, ushering in a distinct approach termed the "prompt-based learning" paradigm, in which LLMs model the probability of text directly (Liu et al., 2023). To make the LLMs perform better on downstream tasks with prompting, the topic of "Prompt Engineering" addresses the interests of researchers and practitioners and achieved some success in recent years. For instance, some prompting techniques like zero- and few-shot prompting (Brown et al., 2020), chain-of-thought prompting (Wei et al., 2022), self-consistency (Wang et al., 2022), generate knowledge prompting (Liu et al., 2021), etc., have been proven to be powerful for many specific tasks.

In summary, the field of AI-aided disaster reconnaissance and SHM still have not fully benefited from the latest NLP technologies, especially powerful LLMs. Besides the structural attributes extracted by well-trained CNN models using image data, a large amount of metadata including text descriptions and geo-information are nowadays collected during the reconnaissance efforts. Therefore, an idea arises to develop a novel multi-modality framework incorporated with CV, NLP, and LLM technologies to fully utilize information from extracted attributes and metadata to automatically summarize key contents and generate rapid reconnaissance reports.

## 3　LLM-DRS Framework

In this section, an LLM-based disaster reconnaissance summarization framework is proposed, namely, LLM-DRS. As illustrated in Figure 1, the LLM-DRS framework mainly consists of three major parts: (1) Planed investigation and data collection, (2) data processing and matching, and (3) reconnaissance report generation.

### 3.1　Reconnaissance Automation Plan

To build a close-to-automated reconnaissance and assessment framework using AI technologies, a standard reconnaissance plan is required. Specifically, the collection of both photographs (images) and the corresponding metadata of structures should follow a well-designed on-site investigation plan to not only ensure both the quantity and quality but also make the process and collected data standardized and structured for further effective processing.

Based on the workflow of the Field Assessment Structural Teams (FASTs) of the Structural Extreme Events Reconnaissance (StEER) Network (Kijewski-Correa et al., 2021), an investigation reconnaissance plan is proposed herein. Once a disaster (e.g., earthquake) takes place, the investigators, equipped with personal protective equipment, cameras, and other professional instruments, will be sent to the affected regions for on-site investigation. Instead of using traditional survey forms or notebooks to record the findings during the investigation, specified data collection software can be adopted, e.g., a Fulcrum APP, developed by StEER (Kijewski-Correa et al., 2021), which contains both vision data and metadata, refer to Figure 2 for a screenshot of the APP. The vision data usually refer to the images or videos taken for the specific damaged structures or components taken by cameras, and the metadata commonly consist of geolocations and descriptions of the target structures or components recorded by text or voice notes. The reconnaissance should follow a certain sequence such as building by building. When investigating a building, all data collection follows a hierarchical rule that starts from the system-level investigation of the outside façade with the overall global situation and then focuses on the detailed investigation through entering the structure component-by-component and floor-by-floor. During this process, the investigators should add meta-information to complement the photographs of certain damaged components. Besides, the other investigators, such as engineers in the Virtual Assessment Structural Teams (VASTs), are usually asked to record general information and provide comments on the event, structure, and project (e.g., the earthquake magnitude, geolocation, address, and type of construction of the structure, etc.), which are collected as general metadata from different web sites and media outlets.

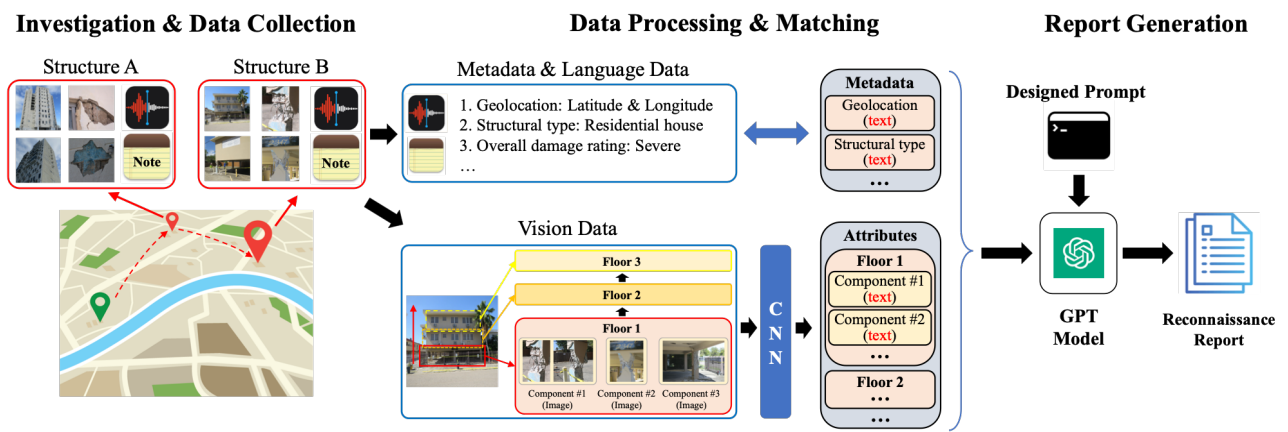

Figure 1. LLM-DRS framework for individual building report generation.

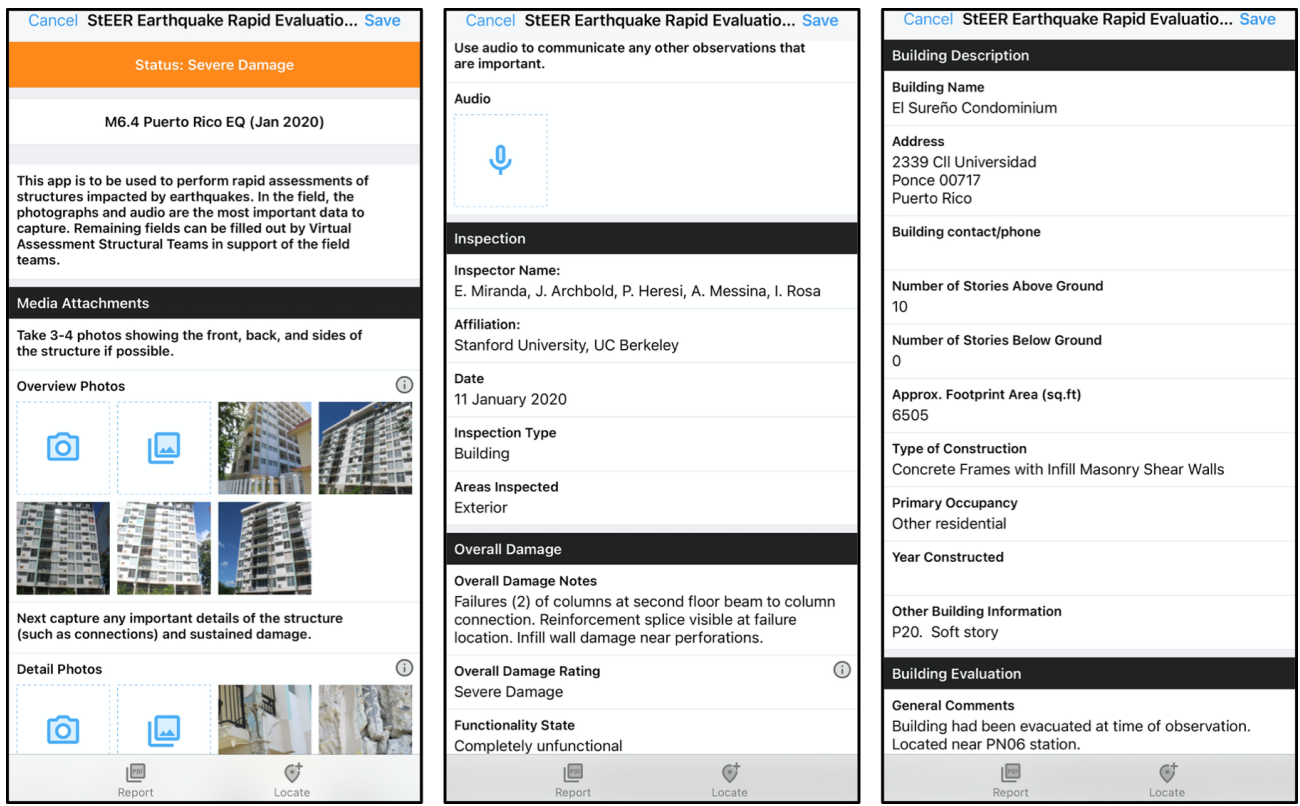

Figure 2. Fulcrum APP screenshot.

### 3.2   Attribute Extraction and Data Matching

Based on the data type, all collected data are separated into two groups, i.e., metadata represented by text and vision data represented by images. In this step, metadata are matched by the attributes of the event and the structures, e.g., time and magnitude of the earthquake, its location, and type of structures, and then reorganized into one file #1 (e.g., JSON, JavaScript Object Notation, which uses human-readable text to store and transmit data objects). Subsequently, the vision data are processed to extract structural attributes and to transform them into a text representation. Based on previous works, key structural attributes can be automatically extracted by well-trained CNN models, e.g., the Structural ImageNet Model (SIM). In this study, the following attributes, defined in the $\phi$-Net, are considered: (1) damage state, (2) spalling condition, (3) material type, (4) collapse mode, (5) component type, (6) damage level, and (7) damage type. Therefore, each image is processed by a SIM to generate its text description of structural attributes following the rules in the $\phi$-Net. Based on the investigation plan defined above, the information of all images is extracted component-by-component and then floor-by-floor. Subsequently, such information is stored in another file #2, reorganized by the floor information. Finally, these two documents (#1 & 2) are merged and further passed to the next step.

### 3.3   LLM-based Report Generation

The above organized textural information is fed to the LLM with a set of carefully crafted prompts to generate the summary report for either an individual structure or the affected regions based on the attributes and metadata from all investigated structures within the region.  For an individual structure, the crafted prompt consists of two parts: (1) system message and (2) user prompt, where the system message is used to prime the model with context, instructions, or other information relevant to the used case, and the user prompt specifies the goal of the report generation. In this study, the system message considers three main points: (i) task description, which defines the role of an expert and describes the details of the ongoing task, (ii) explanations of technical words, which introduce explains the technical words, and (iii) overall rules, which further add format and language constraints. Partially designed prompts are presented as follows:

1. "Act as a domain expert in structural engineering and earthquake engineering working on a project of post-earthquake reconnaissance and structural health condition evaluation."
2. "Metadata: The general information of the event and the structures, e.g., the magnitude of the earthquake, damage state, ..."
3. "Make the tone formal, technical, and professional, and the generated summary in the form of a technical report."

Once the reports of individual structures are generated, an overall summary for the affected region containing these structures can be generated accordingly by feeding all reported into the LLM model. Furthermore, the photographs can be included in the reports of individual structures, and a map with markers representing all investigated buildings can also be included in the report of the affected region. In summary, at the end of the LLM-DRS framework pipeline, key information obtained during the site investigation has been summarized into a series of reports and presented in a more read-friendly and information-dense manner.

## 4   Case Study: 2020 Puerto Rico Earthquake

In this paper, the reconnaissance efforts of the 2020 Puerto Rico earthquake are adopted as a case study. The main earthquake occurred on January 7th, 2020, at 8 km S of Indios at 4:24 am local time with a magnitude of 6.4, followed by thousands of aftershocks. From January 8th to 12th, StEER sent out a team of FAST and conducted field reconnaissance in six cities in Puerto Rico. The team adopted a StEER-developed Fulcrum APP to collect and organize the data including both vision and metadata, where some of them are then used for this case study. More details can be found in (Miranda et al., 2020).

Based on the collected data and the proposed LLM-DRS in Section 3, the reconnaissance summarization reports are generated for each investigated structure and the entire region. For illustration purposes, the results of two structures are presented in this paper. In Figure 3(a), a report is generated for a local three-story residential building, namely, San Jorge Condominium. The short individual structure report successfully covers the basic information such as geolocations and building type. Based on the vision data, the column components and their damage state and spalling condition are accurately identified. More interestingly, the number of damaged columns (4) is also automatically counted through the LLM model, which provides convenience for the user and opens opportunities for further global-scale damage assessment. In addition, the assessment comment provided by the FAST team from the metadata is included in the final part of the report for reference. Similarly, in Figure 3(b), a short report for seven-story building, namely, Calle Salud Ponce, is generated, which indicates a good alignment with input metadata and structural attributes from image data.

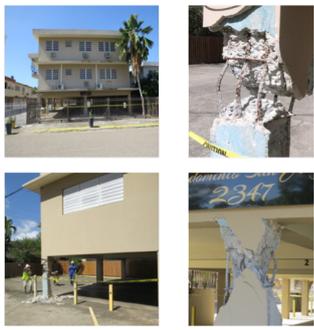

(a) Summarization of the San Jorge Condominium

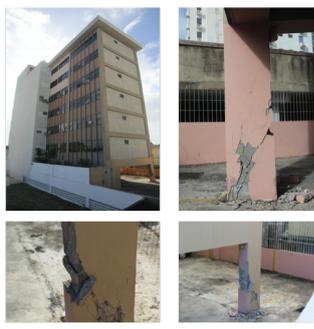

(b) Summarization of the Calle Salud Ponce

*Figure 3. Sample reports generated for two investigated structures in Puerto Rico after 2020 earthquake.*

As an illustrative example of the regional report generation, an overall summary for three structures (including the above two in Figure 3) within a 2 km radius centered at the coordinates (66° 37' 13.44" W, 17° 59' 59.28" N) is generated as shown in Figure 4. More specifically, the second paragraph succinctly outlines the key findings for each of the three structures individually. Notably, paragraphs 3 to 5 expound upon the similarities and characteristics shared by these structures, as automatically generated by the LLM. In the final sentence of the report, the LLM offers an evaluation suggestion by considering the ratings provided by the FAST team for all three structures, providing a concise overview of this affected region.

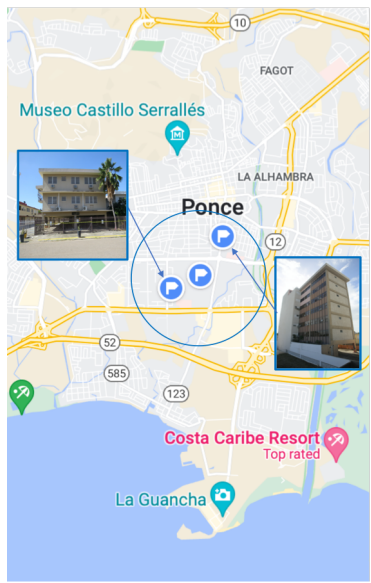

> The region in Puerto Rico, encompassing the coordinates 17.9998, -66.6204 to 18.0074, -66.6125, has been severely affected by the M6.4 earthquake in January 2020. Three structures in this region were inspected and all were found to have severe damage, rendering them mostly to completely unfunctional.
>
> The first structure, a three-story reinforced concrete soft story building named San Jorge Condominium, was found to have multiple failed columns in the soft story. The second structure, a seven-story building constructed with concrete frames and infill masonry shear walls, had two severely damaged columns at the first story. The third structure, a three-story concrete moment frame building named McManus Building at the Catholic University of Puerto Rico, had cracks in south walls and severely cracked columns in shear.
>
> All three structures had heavy damage to their columns, with spalling conditions and combined damage types. The first and third structures also had minor to severe racking damage to walls and other structural damage. The first structure had a moderate level of collapse, partial collapse, or building off foundation, while the other two structures did not have this issue.
>
> The primary occupancy of the first and second structures was residential, while the third structure was a school. All structures had been evacuated at the time of the survey. The inspections were carried out by a team from Stanford University and UC Berkeley, and the data was managed by Mason Moritz and Jorge Archbold.
>
> In conclusion, the region has suffered significant structural damage due to the earthquake, with severe damage to building columns being a common issue. Further detailed evaluation and remedial actions are recommended.

*Figure 4. Sample report with 3 investigated structures generated for the affected region in Puerto Rico after 2020 earthquake.*

## 5  Conclusions

To make full use of multi-modality data collected after a disaster and for the purpose of generating reconnaissance reports for rapid evaluation, a novel LLM-based disaster reconnaissance summarization framework, namely, LLM-DRS, is proposed in this study. It first introduces a standard reconnaissance effort, where the collection of both vision data and the corresponding metadata of structures should follow a well-designed on-site investigation plan. Subsequently, text-based metadata and image-based vision data are further processed and matched to the designated format into one file, making use of the well-trained Deep Convolutional Neural Networks from the PEER Hub ImageNet ($\phi$-Net) to extract the key text-based attributes from these images. Finally, based on crafted prompts, the LLM model, such as GPT-4, generates the summary report of either an individual structure or the affected regions based on the attributes and metadata from all investigated structures within the region. A case study of a reconnaissance project of 2020 Puerto Rico earthquake, where all data were collected by the FAST team of StEER, demonstrates the effectiveness of the proposed LLM-DRS to generate good quality and informative reports for both individual structures and a certain region with multiple investigated structures. In summary, the shown excellent performance of LLM-DRS and its characteristics of reconnaissance report automation reveal the great potential for the field of rapid post-disaster assessment, which is worth further development and illustrations through more case studies.

## 6  Acknowledgments

This research received funding support from: (i) National Science Foundation under Grant No. CMMI 1841667, (ii) California Department of Transportation (Caltrans) for the "Bridge Rapid Assessment Center for Extreme Events (BRACE2)" project, Task Order 001 of the PEER-Bridge Program agreement 65A0774 to the Pacific Earthquake Engineering Research (PEER) Center, (iii) Artificial Intelligence Institute for Food Systems (AIFS), https://aifs.ucdavis.edu/, and (iv) Taisei Chair of Civil Engineering at the University of California, Berkeley. A special thanks to Dr. Jorge Archbold, Universidad del Norte, who collected and provided us the valuable Puerto Rico reconnaissance data through the Fulcrum APP.